\documentclass[letterpaper, 10 pt, journal, twoside]{ieeetran}

\IEEEoverridecommandlockouts                              

\usepackage{times}

\usepackage{amsmath}
\usepackage{cite}
\usepackage[hyphens,spaces,obeyspaces]{url}
\usepackage[hidelinks]{hyperref}

\usepackage{graphicx}
\usepackage{comment}
\usepackage{color}  
\usepackage{xcolor}
\usepackage{amsfonts} 
\usepackage{subfig}

\newcommand{\change}[1]{{#1}}

\title{Learning Spring Mass Locomotion: \\ Guiding Policies with a Reduced-Order Model
}

\author{Kevin Green, Yesh Godse, Jeremy Dao, Ross L. Hatton, Alan Fern, and Jonathan Hurst
\thanks{Manuscript received: October, 14, 2020; Revised January, 17, 2021; Accepted February, 18, 2021.}
\thanks{This paper was recommended for publication by Editor Abderrahmane Kheddar upon evaluation of the Associate Editor and Reviewers' comments.
This work was supported by DARPA contract W911NF-16-1-0002, NSF Grant No. CMMI-1653220 and NSF Grant No. DGE-1314109. Thanks to Stephen Offer and Intel Labs for their computing resources and support.} 
\thanks{All authors are with the Collaborative Robotics and Intelligent Systems Institute, Oregon State University, Corvallis, Oregon 97331. {\tt\footnotesize \{\textbf{greenkev}, godsey, daoje, alan.fern, ross.hatton, jhurst\}@oregonstate.edu}}%
\thanks{Digital Object Identifier (DOI): see top of this page.}
}

\begin{document}

\maketitle


\begin{abstract}
In this paper, we describe an approach to achieve dynamic legged locomotion on physical robots which combines existing methods for control with reinforcement learning. 
Specifically, our goal is a control hierarchy in which highest-level behaviors are planned through reduced-order models, which describe the fundamental physics of legged locomotion, and lower level controllers utilize a learned policy that can bridge the gap between the idealized, simple model and the complex, full order robot.
The high-level planner can use a model of the environment and be task specific, while the low-level learned controller can execute a wide range of motions so that it applies to many different tasks.
In this letter we describe this learned dynamic walking controller and show that a range of walking motions from reduced-order models can be used as the command and primary training signal for learned policies.
The resulting policies do not attempt to naively track the motion (as a traditional trajectory tracking controller would) but instead balance immediate motion tracking with long term stability.
The resulting controller is demonstrated on a human scale, unconstrained, untethered bipedal robot at speeds up to 1.2 m/s.
This letter builds the foundation of a generic, dynamic learned walking controller that can be applied to many different tasks.
\end{abstract}


\IEEEpeerreviewmaketitle

\section{Introduction}

\begin{figure}
    \centering
    \includegraphics[clip,width=0.98\columnwidth]{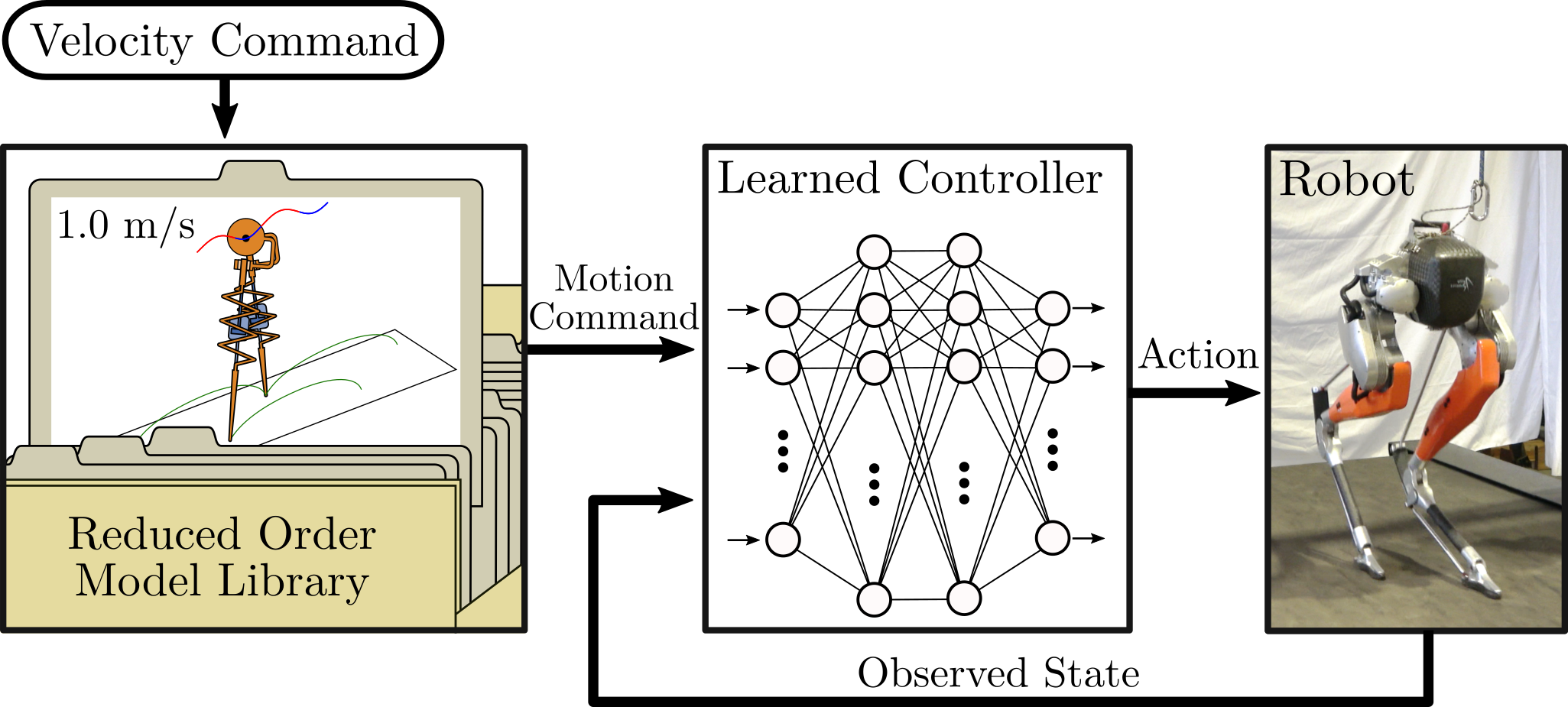}%
    \caption{\change{Our proposed control hierarchy which demonstrates a learned controller for a legged robot that is commanded using reduced-order model motions. In future work, this library of reduced order model motions can be replaced by a dynamic motion planner.}}
    \label{fig:leadDiagram}
\end{figure}

\IEEEPARstart{A}{} powerful approach to control agile and dynamic legged robots is to use a control hierarchy that combines specific domain knowledge of legged locomotion with the power of deep reinforcement learning.
The long term goal is to enable robots to be able to navigate quickly through previously unseen environments with agility that approaches or exceeds that of humans and animals.
The control hierarchy should consist of a low-level walking controller generated through reinforcement learning that can account for and exploit the passive dynamics of the physical robot.
This low-level controller receives motion commands from a terrain-aware motion planner.
The commands from the planner must be rich enough to sufficiently direct the walking controller while still being as simple as possible.
\change{This letter focuses on the learned \emph{controller} and its interface, so we elect to use a library of precomputed motions (Fig. \ref{fig:leadDiagram}).}

Learned controllers have incredible potential to create dynamic locomotion, but to be integrated into a control hierarchy we need an effective control interface. 
Dynamic legged locomotion is by its nature underactuated, hybrid, unstable, nonlinear, and must be able to operate with significant ground uncertainty.
These challenges may be addressed by deep neural networks acting as controllers because of their ability to encode highly nonlinear control policies.
However, if we would like to develop more complex behaviors such as autonomous navigation through unknown, obstacle filled environments, we will need to extend this approach.
It may be possible to expand the learning problem so that the same policy that dynamically controls the robot also interprets the world around it and chooses how to move to the goal, but we choose not to take this approach because of the challenge of generalization to new tasks, sensors and environments.
Instead, we seek to create a learned controller that can be directed by other intelligent systems in a modular hierarchy.
\change{Recent work explores the use of hierarchical learned control structures to quadrupedal locomotion. 
Some methods use both a high-level learned policy and a low-level learned policy \cite{jain2019hierarchical, TsounisDeepGait}. 
Another method combines a low-level model based controller with a high-level, learned gait planner \cite{da2020learning}.}

\begin{figure*}
    \centering
    \includegraphics[width=0.90 \textwidth]{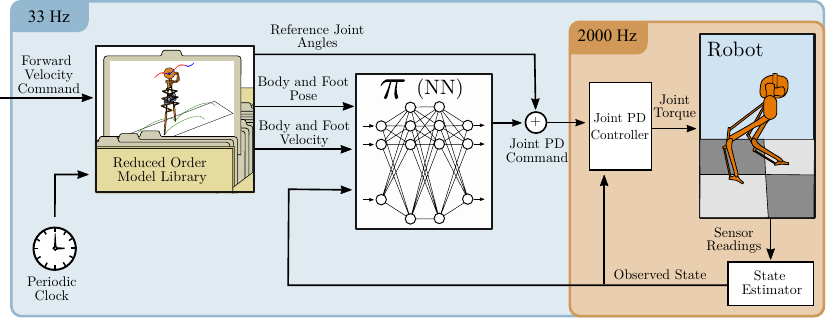}
    \caption{The control diagram for both learning in simulation and running on hardware. At a relatively slow 33 Hz, the library is sampled and the learned policy is evaluated. At a much faster 2000 Hz, the joint PD controller is evaluated, the commanded torques are sent to the motors, and the state estimator is updated. The inputs to the learned policy are only the reduced-order model motion and the state estimate.}
    \label{fig:control_diagram}
\end{figure*}

Reduced-order models of locomotion capture the core dynamics of locomotion which make them a compelling control interface.
The most common models used to plan motion for bipedal robots are inverted pendulum models, which consist of a point mass and massless legs that can apply forces through ground contact \cite{kajita20013d, Motoi2009LIP, Apgar2018}. 
These models describe the underactuation of dynamic walking as well as the discrete choice of foot step locations while remaining simple enough to plan with in real time.
The spring loaded inverted pendulum (SLIP) model is particularly applicable for agile locomotion because \change{with simple, feed-forward policies} it demonstrates strong stabilizing effects \change{\cite{Green2020PlanningForUnexpected, Heim2019Robustness, Hurst2007Policy}}.
Many agile robots closely resemble the SLIP model \cite{Koditschek1991, raibert1986legged} or are designed with SLIP locomotion as a goal \cite{Rezazadeh2015a, Martin2017Deadbeat}, which motivates our choice to use an actuated variation of the SLIP model in this letter.

In this paper we present a method of using reduced-order models of walking to direct high quality, transferable walking controllers and demonstrate its effectiveness on a Cassie series robot from Agility Robotics.
We use the bipedal actuated SLIP model as the reduced-order model of walking to create a library of gaits across different speeds.
The gaits are optimized using a direct collocation method to create energetically optimal walking cycles.
Following these trajectories makes up 70\% of the reinforcement learning reward while 20\% is for foot orientation and 10\% is for smooth actions. 
The resulting controller produces visually natural motions with clear correspondence to the reference trajectories.
These results show that reduced-order model trajectories are useful tools in training and controlling learned walking controllers.

\section{Background}

Reinforcement Learning is a learning framework in which agents learn what actions to take in order to maximize their cumulative future reward. 
Policy gradient methods, such as Proximal Policy Optimization (PPO) \cite{schulman2017proximal}, are a popular choice of reinforcement learning algorithms that have been successfully applied to generate control policies for robotic systems, including legged robots \cite{Xie2018, deepmind_minitaur}.
An important part of using reinforcement learning to solve complex control problems is the design of the reward function evaluated after each agent-environment interaction.
Most applications of reinforcement learning to legged locomotion employ heuristic reward functions to produce walking behavior \cite{hwangbo2019learning, deepmind_minitaur}.
Though this method has been effective, it is often hard to describe a desired motion through objective, generic reward functions.
The definition must be sufficiently detailed to prevent maladaptive policies from learning motions that exploit features of the reward or simulator and do not accomplish the underlying goal when transferred to hardware.
Researchers often use long and complex reward functions to prevent these maladaptive policies; \cite{hwangbo2019learning} uses 8 different reward terms for a single walking task.

A different approach is to use a single expert trajectory as a reference motion \cite{Xie2018}. 
The reward function encourages motions that are close to the specified expert trajectory, which can result in a policy that closely recreates the desired motion.
The use of the reference information discourages exploitative policies since the desired motion is now densely and properly represented.
When a reference motion reward makes up a significant portion of the total reward, it severely restricts the space of solutions to be those near the reference motion.
For some tasks this restriction is unacceptable; however, for bipedal locomotion restricting the final behavior to resemble a normal walking gait is actually preferable because it disincentives strange, exploitative behaviors.

We note that the method from \cite{Xie2018} used a single reference trajectory.
To provide effective information for a variety of speeds, this single trajectory was ``stretched" and ``compressed" to higher and lower speeds, sometimes creating trajectories that were physically infeasible.
These infeasible trajectories may harm the learning problem by making the trajectory matching reward signal conflict with the dynamics of the system. 
Our work mitigates this conflict by using reduced order model trajectories with inverse kinematics to produce feasible walking trajectories for every training speed.
Furthermore, the use of a reduced-order model allows for greater flexibility in the control system, by allowing future work to use the reduced-order model trajectories as a form of higher level planning.

\section{The Control Hierarchy}

We created a control hierarchy (Fig. \ref{fig:control_diagram}) to allow us to train and test a walking controller that utilized reduced-order model motion.
The only external input to the system is a human operator's forward velocity command.
Internally, a periodic clock increments forward through the walking cycle.
The velocity command and clock are inputs to a library of reduced-order model motions (\S \ref{sec:ROM_Libaray}).
The library returns the positions and velocities of the reduced-order model's body and feet for use as input to the learned policy.
Additionally, the library contains a set of robot-specific motor angles that correspond to a robot pose that match the body and foot positions.
The learned policy is evaluated and the output is summed together with the reference joint angles to form the motor proportional-derivative (PD) command.
These commanded angles are sent to the high frequency control loop (\S \ref{sec:high_freq}).
This control loop evaluates the PD controller, sends torques to the motors, measures robot sensors, filters sensor data, and estimates the full state of the robot.
This structure is utilized not just in hardware but also in the simulation environment we use to train the learned policy ($\pi$).

\subsection{The Reduced-Order Model Library}
\label{sec:ROM_Libaray}
Our motion library consists of \change{task space (body and foot)} trajectories for periodic walking over a range of forward speeds.
Body and foot trajectories will be used as an input to the learned policy as well as the majority of the reward signal.
\change{These trajectories will not be strictly dynamically feasible on the full order robot, but they should describe a nearly feasible center of mass motion that can be produced by valid ground reaction forces at the feet.}
To create each trajectory we optimize the reduced-order model and augment it with a minimum-jerk swing leg profile.
We additionally calculate motor angles for each motion through offline inverse kinematics to use as a feed forward term into the PD controller.

\begin{figure}
    \centering
    \includegraphics[width=0.5\columnwidth]{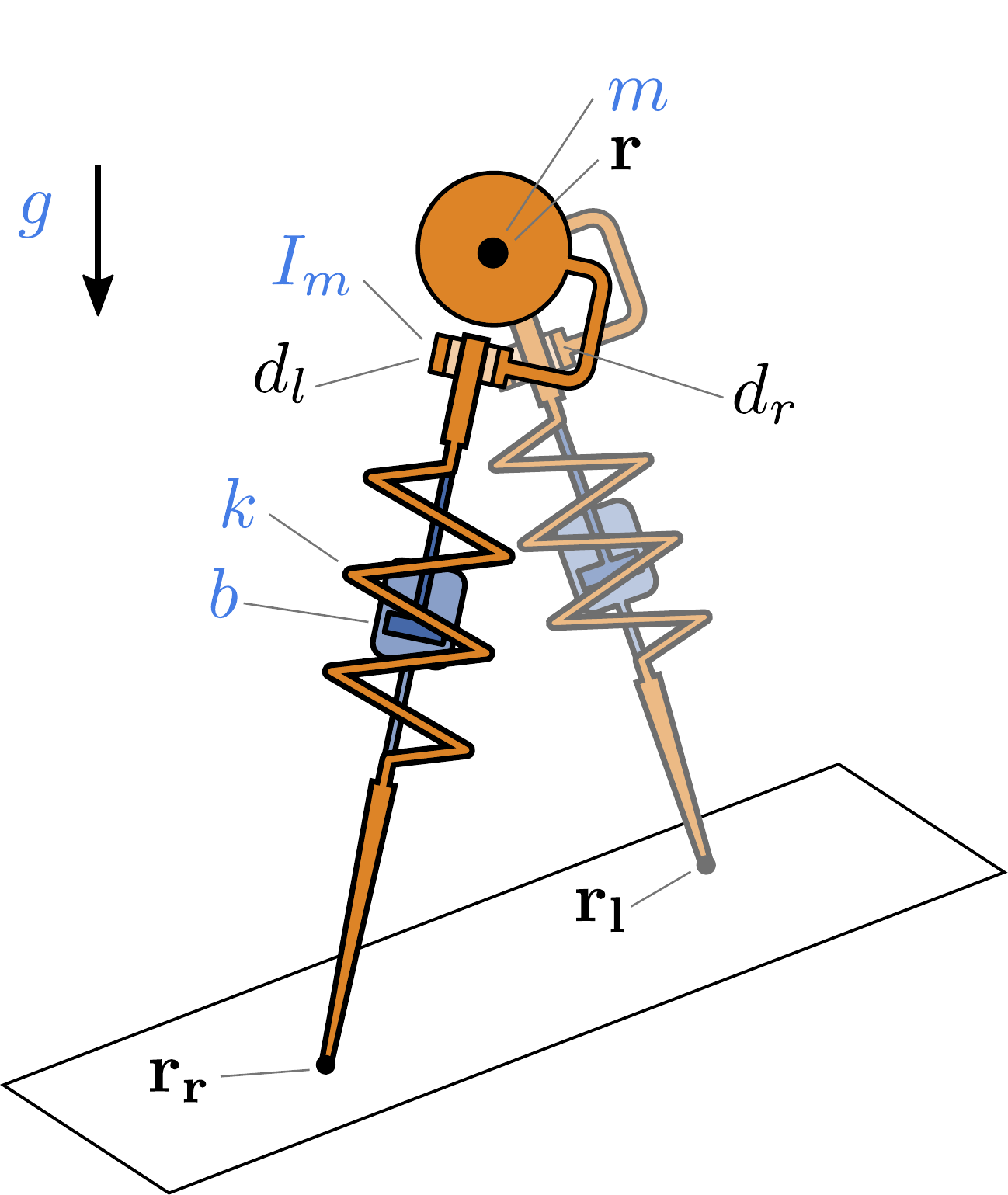}
    \caption{The 3D, bipedal actuated spring loaded inverted pendulum model which we optimize to create walking gaits for our motion library. This model has a point mass body with no rotational inertia and two massless, compliant, actuated legs.}
    \label{fig:ASLIP_Labeled}
\end{figure}

\begin{table}
  \begin{tabular}{ l  c  l    l }
                    & Symbol & Value and Unit & Description \\ \hline
    States          & $\mathbf{r}$         & -        [$m$,$m$,$m$]                 & 3D body position\\
                    & $d_l$       & -        [$m$]               & Left leg actuator setpoint\\
                    & $d_r$         & -        [$m$]                 & Right leg actuator setpoint\\
    \hline
    Parameters      
                    & $m$         & $30$       [$kg$]         & Body mass\\
                    & $k$         & $3000$       [$N/m$]         & Leg spring stiffness\\
                    & $b$         & $2$     [$Ns/m$]    & Leg spring damping\\
                    & $I_m$         & $10$       [$kg$]         & Leg actuator linear inertia\\
                    & $g$         & $9.81$     [$m/s$]    & Gravitational acceleration\\
    \hline
    Continuous          
                    &$u_l$         & -        [$m/s^2$]                 & Left actuator acceleration\\
      Inputs              &$u_r$         & -        [$m/s^2$]                 & Right actuator acceleration\\
    \hline
    Discrete           
                    &$\mathbf{r_l}$         & -        [$m$,$m$,$m$]                 & Left foot stance position\\
     Inputs               &$\mathbf{r_r}$         & -        [$m$,$m$,$m$]                 & Right foot stance position\\
    
  \end{tabular}
  \caption{States, parameters and control inputs for the actuated SLIP model used to generate the library of motions.} \label{tab:aSLIPparameters}
\end{table}

The motions in the library are energetically optimal periodic gaits of the 3D, bipedal actuated SLIP model.
This model consists of a point mass body and two massless legs (Fig. \ref{fig:ASLIP_Labeled}).
Parameters of the model were chosen to closely resemble the Cassie robot, see Table \ref{tab:aSLIPparameters}.
Each leg has a extensible actuator in series with a spring and a damper.
The trajectory optimization method we use is a direct collocation method where the state and inputs are discretized and dynamics are enforced through equality constraints between sequential states \cite{posa2013direct}.
Our implementation is not contact-invariant so we needed to specify the contact sequence.
We define a walking contact sequence made up of alternating single stance and double stance phases.
We vary the average forward velocities from 0 to 2 m/s in steps of 0.1 m/s as an equality constraint on the final state.
The energetic objective function we use is
\begin{equation}
    f = \int_0^T (u_l^2 + u_r^2)dt
\end{equation}
where $u_l$ and $u_r$ are the accelerations of the leg actuators.
This cost function represents the resistive losses in an electric motor when it applies forces to accelerate the inertia of its rotors. 
\change{We ensure kinematic feasibility by including a conservative constraint on the maximum and minimum leg extension.}
To generate the swing foot motions we calculated minimum integrated jerk squared motion profiles.
These profiles connect the footholds from the trajectory optimization with a specified vertical clearance height at the midpoint.
\change{We use a modified version of COALESCE \cite{Jones2014} to generate the problem and its analytical gradients and IPOPT \cite{wachter2006implementation} to solve the problem.}

\begin{figure}
    \centering
    \includegraphics[width=0.80\columnwidth]{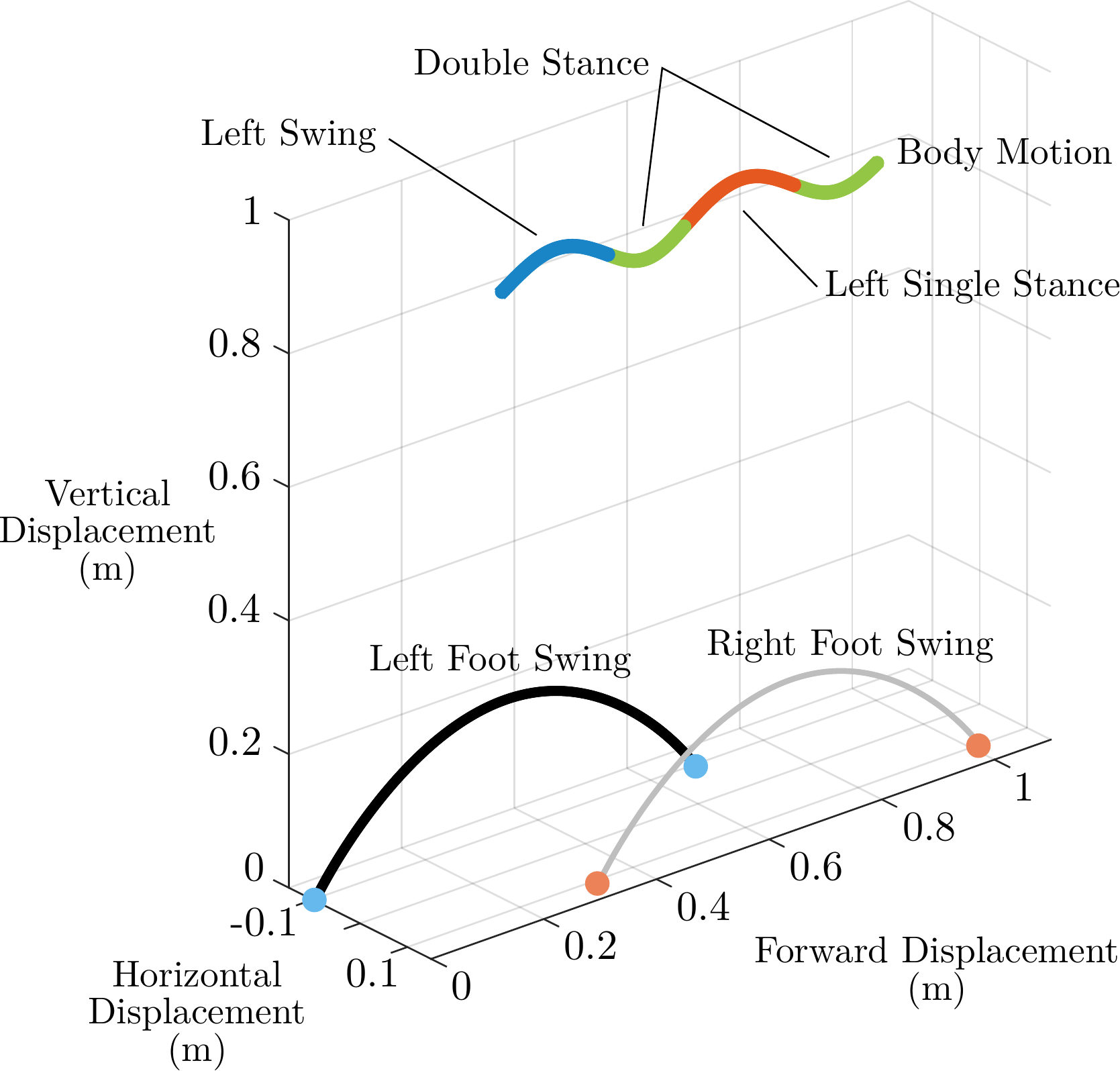}
    \caption{The optimized task space motion of the reduced-order model for 1.0 m/s walking. The hybrid phase is shown through the color of the body trajectory. This motion together with its velocities make up the motion library.}
    \label{fig:opt_ASLIP_Motion}
\end{figure}

An example optimal motion with a mean velocity of 1.0 m/s is shown in Fig. \ref{fig:opt_ASLIP_Motion}.
The different phases can be seen in the main body's path where it reaches its peak height in single stance and the minimum height in double stance.
The feet follow smooth paths with the 0.2 m specified vertical clearance height.

\subsection{High Frequency Control Loop}
\label{sec:high_freq}
The task space motions of the reduced-order model are inputted to a learned policy running at 33 Hz which outputs delta motor positions $\delta_a$.
These deltas are added to the baseline joint angles from the library, $\hat{a}$, to create the final command for the PD controller, $a = \hat{a} + \delta_a$.
The use of joint angle baseline actions was used in prior work on learned walking controllers for Cassie \cite{XieCORL}.
The actions are passed to a joint level PD controller with fixed feedback gains running at 2 kHz.
Note that the learned policy only outputs position targets while the velocity targets are always set to zero. This means the proportional-derivative controller should more accurately be called a proportional-damping controller.
We choose this structure as previous work has shown learning PD targets to be easier and produces higher quality motions \cite{Peng_action_space}.

\section{Reinforcement Learning} 

The optimal learned controller should be able to capture the most important features of the reduced-order model's motion and translate them into a stable walking behavior that functions on hardware.

\subsection{Problem Formulation}

The inputs to the policy are the estimated robot state and the desired positions and velocities of the robot's pelvis and feet.
The estimated robot state contains the pelvis height, orientation, translational velocity, rotational velocity, and translational acceleration in addition to the positions and velocities of the actuated and unactuated joints on Cassie.
This state estimate together with the commanded pose and velocity from the reduced-order model library form a $64D$ input space. 
The output of the learned policy is a $10D$ vector containing motor position targets for each of Cassie's actuated joints.

We learn control policies by using a simulated model of Cassie\footnote{Simulation and state estimation library available at \url{https://github.com/osudrl/cassie-mujoco-sim}} in the MuJoCo Physics simulator \cite{mujoco}. 
Our dynamic model of Cassie includes the reflected inertia of each motor (defined as ``armature" in MuJoCo).
We also attempt to model actuator delay by limiting when new torque commands are actually executed.
Desired torques only take effect \change{0.003 seconds (6 time steps of the high frequency, PD control loop)} after being ``sent" to the simulator.
We believe these extra facets of the model help improve the policy's robustness against differences in simulation and reality, enabling cleaner sim-to-real transfer. 

An important part of this setup is that even during training we use an \textit{estimate} of the state rather than the true state.
Though we have access to the true simulated state we instead pass the simulated sensor values into a state estimator to get a simulated ``observed state."
This incorporates simulated sensor noise and state estimator dynamics into the learning process, which is an essential part of making policies robust enough for sim-to-real transfer.
This allows us to use the exact same controller structure on hardware, effectively just switching out the simulated robot for the real robot.

\subsection{Learning Procedure}

\begin{table}
\vspace{4pt}
\begin{center}
  \begin{tabular}{ l | l }
    Parameter  & Value \\ \hline
    Adam learning rate  & $1\times 10^{-4}$ \\
    Adam epsilon & $1 \times 10^{-5}$ \\
    discount ($\gamma$) & $0.99$ \\
    clipping parameter ($\epsilon$) & $0.2$ \\
    epochs & $3$\\
    minibatch size & $64$\\
    sample size & $5096$ \\
  \end{tabular}
\end{center}
\caption{PPO Hyperparameters}
\label{tab:PPOParameters}
\end{table}

The reinforcement learning algorithm we use is an implementation of PPO\change{\footnote{Reinforcement Learning code, reward functions, and ASLIP reference motions available at \url{https://github.com/osudrl/ASLIP-RL}}} with parallelized experience collection and input normalization\cite{schulman2017proximal}.
Our policy is a fully connected feed-forward neural network with 2 hidden layers of 256 nodes each.
We choose to use fixed covariance instead of making it an additional output of the policy.
The hidden layers use the ReLU activation function and the output is unbounded.
\change{This architecture was chosen because previous work found it was large and deep enough to generate high quality locomotion across a range of walking speeds \cite{Xie2018}.}
More information on training hyperparameters can be found in table \ref{tab:PPOParameters}.
At the start of each episode, a reference trajectory is randomly selected from the reduced-order model library and the simulated model of Cassie is set to a random starting position in the trajectory's walk cycle.
A single step of agent-environment interaction includes the policy computing an action, sending it to the low-level PD controller which simulates forward 1/33 of a second, and retrieving the next state in the 33 Hz execution cycle.
We define the maximum episode length for this MDP problem to be 400 steps of agent-environment interaction, which corresponds to 12 seconds.
Episodes are terminated when the number of steps reaches the maximum episode length or when the reward for the current step is less than 0.3.
\change{This termination condition encapsulates when the robot falls to the ground or if it deviates excessively from the goal behavior.}

We design the following reward function that is evaluated after each step:
\begin{subequations}
  \begin{align*}
      r = \quad & \: 0.3 \> r_{\text{CoM\_vel}} \>+ 0.3 \> r_{\text{foot\_pos}} \>+ 0.1 \> r_{\text{straight\_diff}} \\
       \>+ & \: 0.2 r_{\text{foot\_orient}} \>+ 0.1 r_{\text{action\_diff}}
  \end{align*}
\end{subequations}
All of the terms in the reward function are computed as the negative exponential of a distance \change{metric.}
This lets us limit the maximum reward per step to 1 and per episode to 400. 

\begin{figure*}
    \centering
    \includegraphics[width=0.75\textwidth]{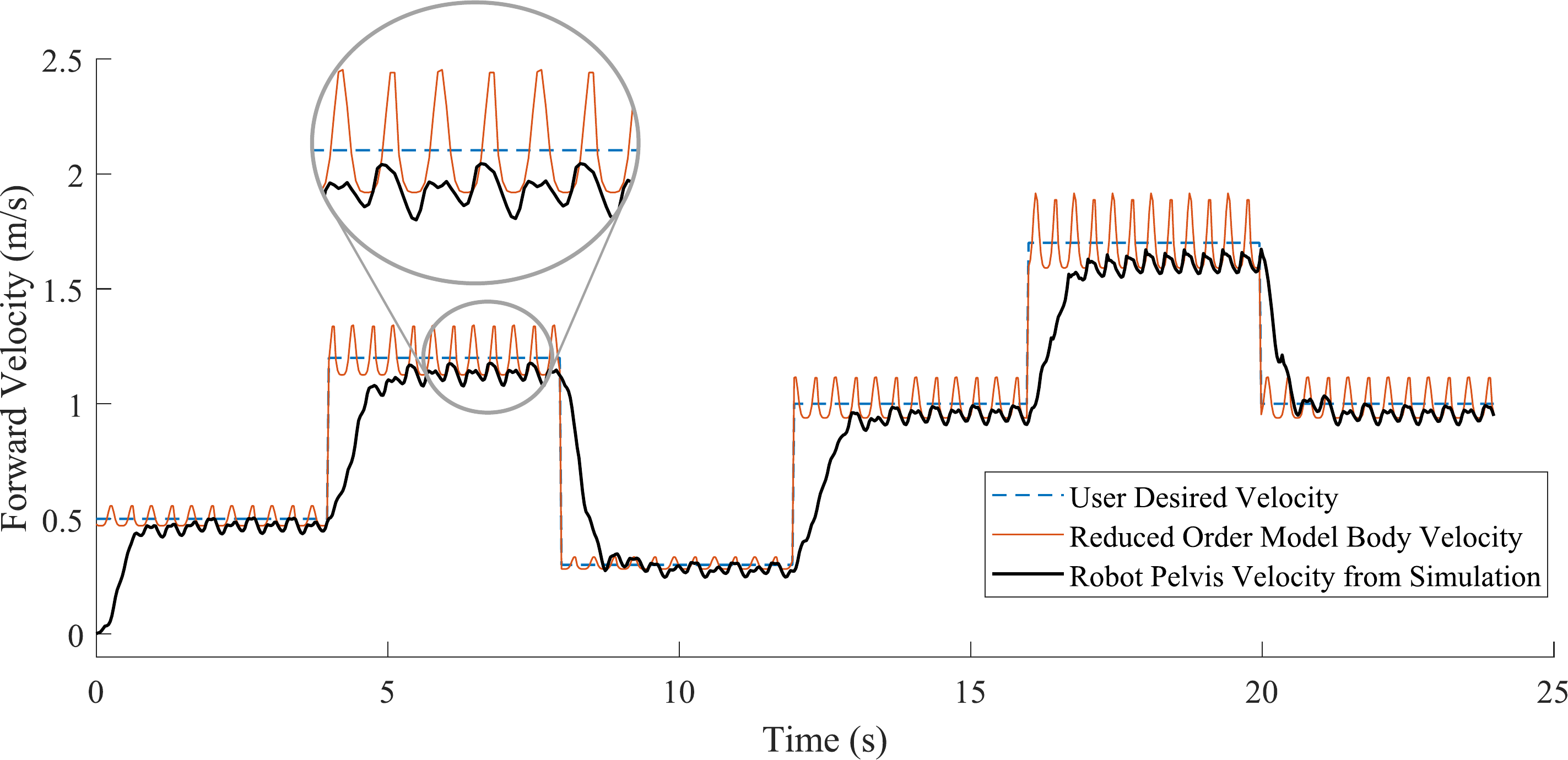}
    \caption{Forward velocity comparison of the user desired velocity, the corresponding reduced-order model center of mass velocity, and the simulated robot's pelvis velocity. As can be seen, the policy closely tracks the \change{user's} desired velocity. \change{The difference between the reduced-order model and robot's velocity show the learned policy does not emulate the spikes in reduced-order model's body velocity which occur at touchdown.}}
    \label{fig:velocity_tracking}
\end{figure*}

The first three terms of the reward function account for 70\% of the total reward.
Together they penalize the differences between the current state of the robot and reference task space position and velocity.
The center of mass velocity matching reward, defined as 
\begin{equation}
r_{\text{CoM\_vel}} = \exp(-||\mathbf{v}_{\text{CoM}} - \mathbf{v}_{\text{refCoM}}||),
\end{equation}
incentivizes matching the robot's center of mass (pelvis) velocity to the reference velocity.
However, $r_{\text{CoM\_vel}}$ is calculated using the \textit{local} pelvis frame which prevents the policy from receiving a large reward for sidestepping or walking diagonally.
To ensure the robot locomotes in the forward direction, $r_{\text{straight\_diff}}$ rewards the lateral robot position being close to zero.

To get the controller to track the reference foot positions, $r_{\text{foot\_pos}}$ rewards the robot's foot positions to be close to the reference motion's foot positions, where the foot position is defined as the position of the foot relative to the body.
On the full order robot, the orientation of the foot joints is important for stable walking on hardware.
However, our reduced order model has point feet which do not describe or incentivize any particular foot orientation.
\change{We reward forward pointing toes and feet parallel to the ground through $r_{\text{foot\_orient}}$.}

A reward term that helps the transfer to hardware is $r_{\text{action\_diff}}$, which penalizes the distance between the last action and the current action and results in smoother action outputs.
Without this term the policy can converge to behaviors that rapidly oscillate the commanded motor angles which is not conducive to success on hardware.

It is important to note that with the inclusion of the $r_{\text{action\_diff}}$ term, reaching the maximum reward is impossible and not an expectation, as the policy would need to output a constant motor angle while tracking the reference motion.
Furthermore, perfectly matching the positions of the reduced-order model at each step is likely not possible because of the significantly more complex dynamics of the full order robot. 
As seen in \cite{Rezazadeh2015a, Martin2017Deadbeat}, directly applying spring-mass behavior to a robot is challenging and sensitive. 
Thus our learned controller should use the spring mass model as a guide towards highly effective walking solutions that work for the robot on hardware.

\section{Results} 

We trained ten different policies from randomized initial weight seeds for our method.
The training process takes just under five hours of wall clock time using 50 cores on a dual Intel$^{\tiny{\text{\textregistered}}}$ Xeon$^{\tiny{\text{\textregistered}}}$ Platinum 8280 server.
The policy learns to step in place after about 25 million timesteps, and converges to a reward of almost 300 after about 175 million timesteps, where it is able to track all of the walk cycles in the reduced-order model library.

\subsection{Simulation} 

\begin{figure}
    \centering
    \includegraphics[width=0.98\columnwidth]{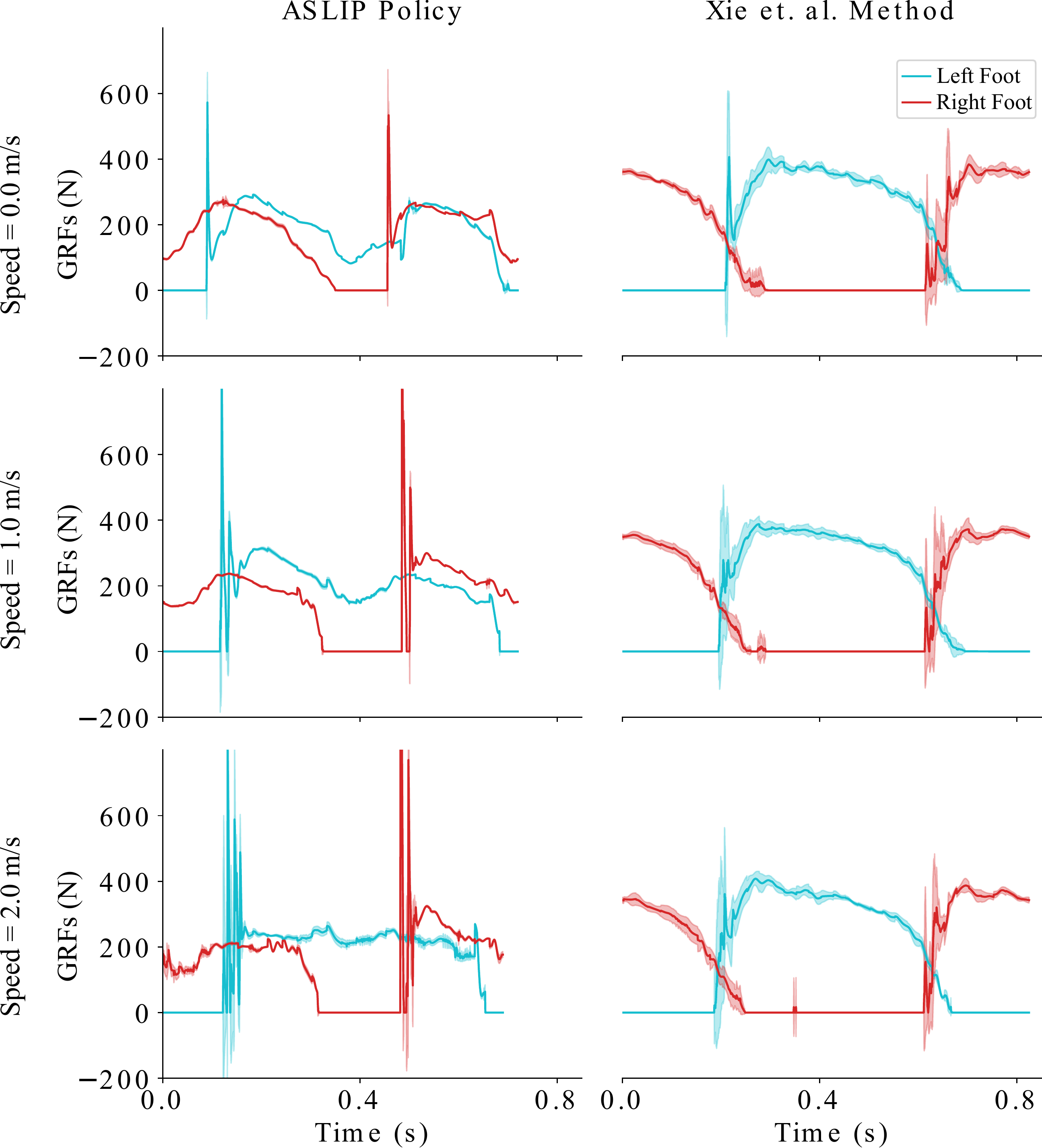} 
    \caption{Ground reaction force profiles of the actuated SLIP policy compared to a single reference trajectory policy for \cite{Xie2018} method. The actuated SLIP policy shows double hump ground reaction forces as is expected from spring mass walking, where the single reference policy shows flattened single hump ground reaction forces similar to linear inverted pendulum walking.}
    \label{fig:GRFs}
\end{figure}

By varying the user-provided forward velocity to the reduced-order model library, we demonstrate the learned control policy's ability to smoothly transition between discrete reference walk cycles (Fig. \ref{fig:velocity_tracking}).
The policy produces walking behavior with oscillations in pelvis velocity that correspond with the oscillations in center of mass velocity from the reduced-order model. 
Furthermore, this velocity tracking succeeds across the broad set of commanded speeds and the transitions between them.

\begin{figure}
    \centering
    \includegraphics[width=0.90\columnwidth]{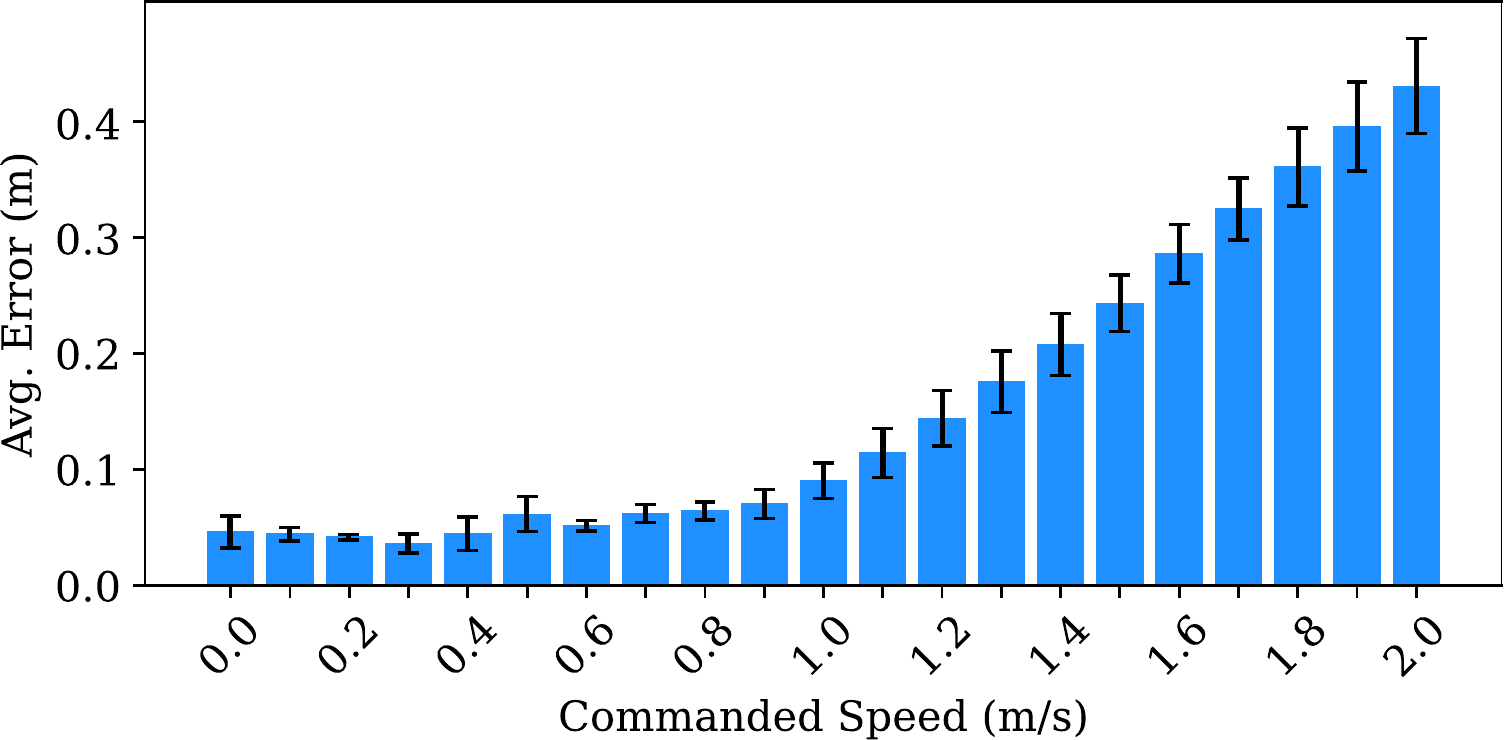}
    \caption{Average foot placement error and standard deviation over 15 foot steps at various speeds. The average error across speeds under 1.0 m/s is relatively low and is sufficient for planning precise footstep locations.
    }
    \label{fig:footpos_err}
\end{figure}

In order to quantify how well desired foot locations can be realized through this control hierarchy, we measure the average error between the foot touchdown locations of the reduced order model commands and the robot in simulation (Fig. \ref{fig:footpos_err}).
At all speeds, we observe that the robot places its feet slightly wider than the reduced-order model.
Above 1.0 m/s, we see that the robot's footsteps lag more and more which corresponds to error in tracking the reduced-order model's forward velocity.
At speeds under 1.0 m/s this placement error doesn't exceed an average of 10 cm.

The ground reaction forces show that our policy produces features indicative of spring mass walking \change{ that were not explicitly incentivized by the reward function}.
We compare the ground reaction forces across different commanded speeds to those from the single reference trajectory policy (Fig. \ref{fig:GRFs}).
Particularly at 0 and 1.0 m/s the actuated SLIP policy has a double hump ground reaction force which is present in spring mass walking.
In comparison\change{,} the single reference trajectory has a single hump ground reaction force with a flattened peak for all speeds.
This type of ground reaction force is expected from a bipedal walking policy that holds the body at a constant height, similar to the linear inverted pendulum \cite{kajita20013d}.
\change{ These ground reaction forces are not themselves a useful measure of performance of the gait, but instead provide us evidence that our learned controller is emulating the dynamics of our reduced order model.}

\subsection{Hardware} 

\begin{figure*}
    \centering
    \includegraphics[width=.75\textwidth]{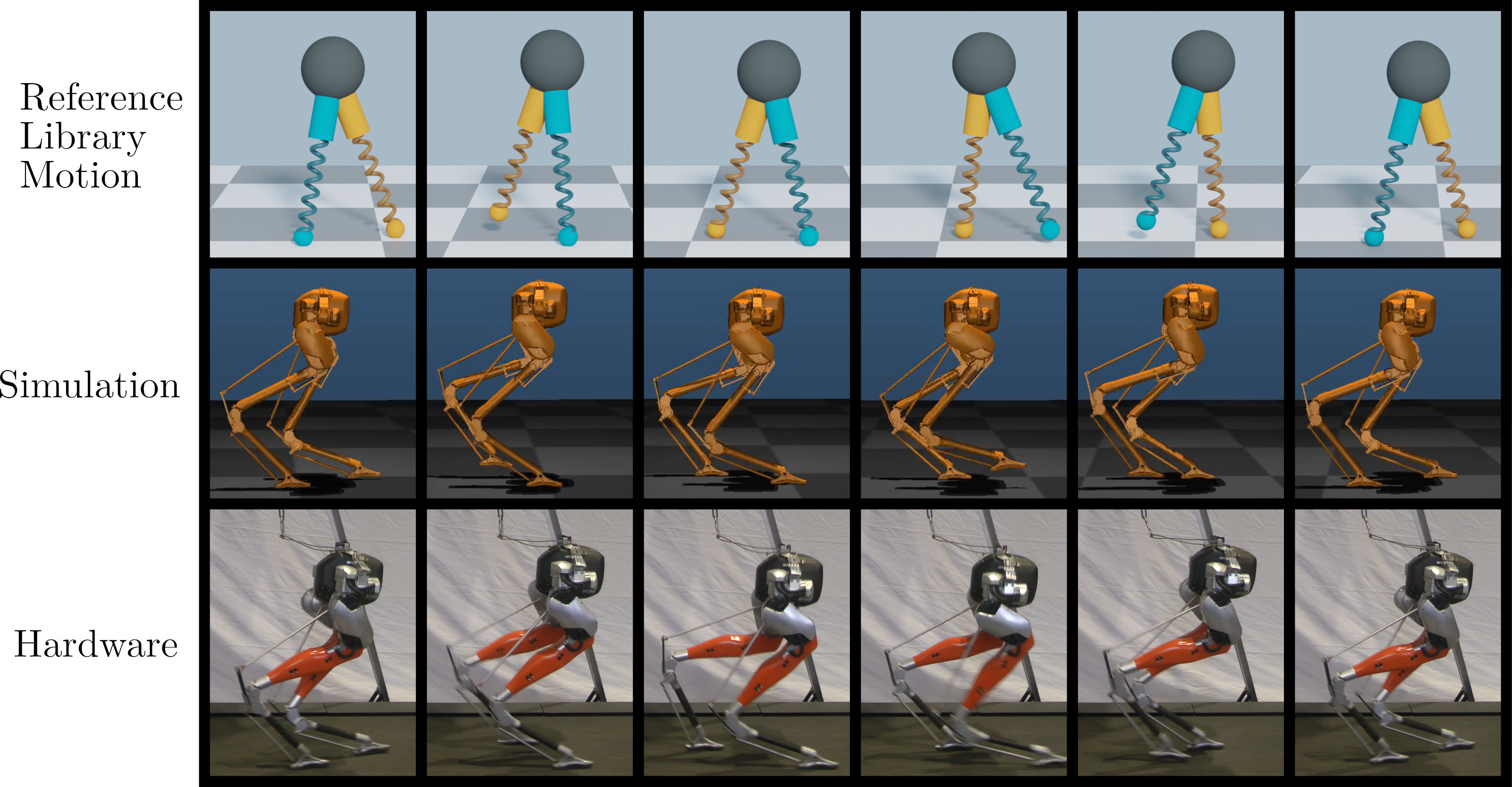}
    \caption{Motion comparison of the reduced-order model, simulation, and hardware \change{at different phases in the gait.} The top images represent the desired reference motion to recreate, the middle images show the learned policy in simulation, and the bottom images show the learned policy executed on hardware. } 
    \label{fig:filmStrip}
\end{figure*}

We directly transfer the policies trained in simulation to hardware, demonstrating that this approach can achieve a strong sim-to-real transfer.
We observe that the the learned walking motion is springy, with slight oscillations in the pelvis velocity and changes in leg length directly corresponding to the same variations in the motion of the reduced-order model. 
This motion correspondence can be seen in Fig. \ref{fig:filmStrip} and our accompanying video, which shows Cassie walking using our learned policy for an extended period of time, as well as a comparison between the motions of the reduced-order model, the learned controller in simulation, and the learned controller on hardware.
The video also shows that the stepping frequency of all three stages in the control hierarchy: reduced-order model, simulation, and hardware, match for the same forward velocity commands.

To test the ability of these policies to rapidly change speeds in hardware, the human operator sends the robot sudden changes in velocity commands.
The results of this trial are shown toward the end in the attached video.
This controller is able to walk significantly faster and with a longer stride than was possible using previous model-based control methods on the same robot \cite{Apgar2018}.

\section{Conclusion} 
\label{sec:conclusion}

In this letter, we have presented an effective control structure for producing spring mass-like motion on a human scale bipedal robot.
This method employs reduced-order model reference trajectories to inform the learning process of the desired task space motion.
We find that this method is successful in producing similar motion to the \change{actuated} SLIP model and generates policies that can realize this behavior on the bipedal robot Cassie.
\change{We found success using the actuated SLIP model as the reduced order model to guide Cassie.
One should consider carefully the choice of reduced order model when applying this work to other robots.}
Continuations of this work will focus on extending the variety of motions the low-level policy can track and improving the foot step location tracking.

This low-level controller will enable many different opportunities for integration of high-level motion planners.
Now that we have a policy capable of following a desired reduced-order model motion, we can work to extend this to generate policies that follow any arbitrary reduced-order model trajectory.
This would allow for incorporating a high-level planner in the reduced-order model space, such as the planner proposed in \cite{clary2018monte}.
Allowing for reactive planning decisions like navigation and obstacle avoidance to happen at the reduced-order model level will also make it significantly easier to achieve fully autonomous agile legged robots.

\section*{Acknowledgments}
We thank John Warila and Dylan Albertazzi for their assistance rendering videos, 
Helei Duan, Jonah Siekmann, Lorzeno Bermillo, and Pedro Morais for productive discussions.

\bibliographystyle{IEEEtran}
\bibliography{references}

\end{document}